\documentclass[runningheads,a4paper]{llncs}

\usepackage{amssymb}
\usepackage{graphicx}

\usepackage{url}
\urldef{\mailsa}\path|{alfred.hofmann, ursula.barth, ingrid.haas, frank.holzwarth,|
\urldef{\mailsb}\path|anna.kramer, leonie.kunz, christine.reiss, nicole.sator,|
\urldef{\mailsc}\path|erika.siebert-cole, peter.strasser, lncs}@springer.com|    
\newcommand{\keywords}[1]{\par\addvspace\baselineskip
\noindent\keywordname\enspace\ignorespaces#1}
\usepackage[utf8]{inputenc} 
\usepackage[T1]{fontenc}    
\usepackage{url}            
\usepackage{booktabs}       
\usepackage{amsfonts}       
\usepackage{nicefrac}       
\usepackage{microtype}      

\usepackage{amsmath,amssymb} 
\usepackage{color}
\usepackage{amsmath}
\usepackage{amssymb}
\usepackage{algorithm}
\usepackage{algorithmicx}
\usepackage{algpseudocode}
\usepackage{algpascal}

\usepackage{mathrsfs}
\usepackage{soul}
\usepackage{multirow}
\usepackage{comment}
\usepackage{xfrac}
\usepackage{rotating}
\usepackage{hhline}
\usepackage[table]{xcolor}
\usepackage{mathtools}
\usepackage{wrapfig}
\usepackage{svg}
\usepackage{xkeyval} 

\usepackage{array}
\newcommand{\PreserveBackslash}[1]{\let\temp=\\#1\let\\=\temp}
\newcolumntype{C}[1]{>{\PreserveBackslash\centering}p{#1}}
\newcolumntype{R}[1]{>{\PreserveBackslash\raggedleft}p{#1}}
\newcolumntype{L}[1]{>{\PreserveBackslash\raggedright}p{#1}}

\usepackage{pgfplots}
\usepackage{pgfplotstable}
\usepackage{filecontents,pgfplots}
\usepackage{tikz}
\usetikzlibrary{arrows,calc,shapes,snakes,positioning,matrix,arrows,decorations.pathmorphing,decorations.text}
\usepackage{sci}
\usepackage{dsfont}
\usepackage{bbm} 
\usepackage{caption}
\DeclareCaptionFont{9pt}{\fontsize{9pt}{10pt}\selectfont}
\begin{document}
\mainmatter  

\title{Vision-based Estimation of \texttt{MDS-UPDRS} Gait Scores for Assessing Parkinson's Disease Motor Severity
}

\titlerunning{Vision-based Estimation of MDS-UPDRS Gait Scores}

\authorrunning{M. Lu et al.}  
 \author{
   Mandy Lu\inst{1}
   \and Kathleen Poston\inst{2}
   \and Adolf Pfefferbaum\inst{2,3}
   \and Edith V. Sullivan\inst{2}
   \and \\ Li Fei-Fei\inst{1}
   \and Kilian M. Pohl\inst{2,3}
   \and Juan Carlos Niebles\inst{1}
 	\and Ehsan Adeli\inst{1,2}
 }

\institute{Computer Science Department, Stanford University, Stanford, CA, USA \and School of Medicine, Stanford University, Stanford, CA, USA \and Center of Health Sciences, SRI International, Menlo Park, CA, USA}

\toctitle{Vision-based Estimation of MDS-UPDRS Gait Scores for Assessing Parkinson's Disease Motor Severity}
\tocauthor{Mandy Lu,  Kathleen Poston, Adolf Pfefferbaum, Edith V. Sullivan, Li Fei-Fei, Kilian M. Pohl, Juan Carlos Niebles, Ehsan Adeli}


\toctitle{Vision-based Estimation of MDS-UPDRS Gait Scores for Assessing Parkinson's Disease Progression}
\tocauthor{Authors' Instructions}
\maketitle

\begin{abstract}
Parkinson's disease (PD) is a progressive neurological disorder primarily affecting motor function resulting in tremor at rest, rigidity, bradykinesia, and postural instability. The physical severity of PD impairments can be quantified through the Movement Disorder Society Unified Parkinson's Disease Rating Scale (\texttt{MDS-UPDRS}), a widely used clinical rating scale. Accurate and quantitative assessment of disease progression is critical to developing a treatment that slows or stops further advancement of the disease. Prior work has mainly focused on dopamine transport  neuroimaging for diagnosis or costly and intrusive wearables evaluating motor impairments. For the first time, we propose a computer vision-based model that observes non-intrusive video recordings of individuals, extracts their 3D body skeletons, tracks them through time, and classifies the movements according to the \texttt{MDS-UPDRS} gait scores. Experimental results show that our proposed method performs significantly better than chance and competing methods with an $F_1$-score of 0.83 and a balanced accuracy of 81\%. This is the first benchmark for classifying PD patients based on \texttt{MDS-UPDRS} gait severity and could be an objective biomarker for disease severity. Our work demonstrates how computer-assisted technologies can be used to non-intrusively monitor patients and their motor impairments. The code is available at {\small \url{https://github.com/mlu355/PD-Motor-Severity-Estimation}}.

\end{abstract}

\section{Introduction}

Parkinson's disease (PD) is a progressive neurological disorder that primarily affects motor function. Early, accurate diagnosis and objective measures of disease severity are crucial for development of personalized treatment plans aimed to slow or stop continual advancement of the disease \cite{venuto2016review}. Prior works aiming to objectively assess PD severity or progression are either based on neuroimages \cite{adeli2016joint,bharti2019neuroimaging} or largely rely on quantifying motor impairments via wearable sensors that are expensive, unwieldy, and sometimes intrusive \cite{hobert2019progressive,hssayeni2019wearable}. With the rapid development of deep learning, video-based technologies now offer non-intrusive and scalable ways of quantifying human movements \cite{kanazawa2018end,chiu2019action}, yet to be applied to clinical applications such as PD. 

\begin{figure}[t]
  \begin{minipage}[c]{0.59\textwidth}

    \includegraphics[width=\textwidth]{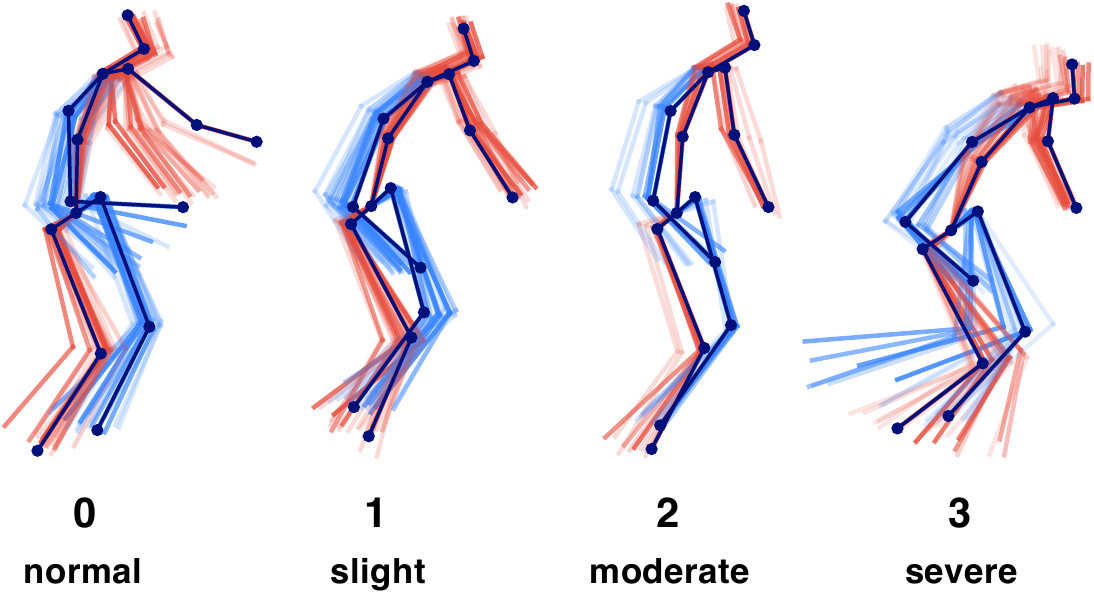}

  \end{minipage}\hfill
  \begin{minipage}[c]{0.38\textwidth}
    \caption{Progressive PD impairments demonstrated by 3D gait (poses fade over time; left/right distinguished by color) with \texttt{MDS-UPDRS} gait score shown below each skeleton. Participants are taken from our clinical dataset. Classes 0 to 2 progressively decrease in mobility with reduced arm swing and range of pedal motion (\ie reduced stride amplitude and footlift) while class 3 becomes imbalanced. } 
    \label{fig:pose_progression}
  \end{minipage}
  \vspace{-9pt}
\end{figure}

PD commonly causes slowing of movement, called bradykinesia, and stiffness, called rigidity, that is visible during the gait and general posture of patients.  The Movement Disorder Society-Unified Parkinson's Disease Rating Scale (\texttt{MDS-UPDRS}) \cite{goetz2008movement} is the most commonly used method in clinical and research to assess the severity of these motor symptoms. Specifically, the \texttt{MDS-UPDRS} gait test requires a subject to walk approximately 10 meters away from and toward an examiner. Trained specialists assess the subject's posture with respect to movement and balance (\eg `stride amplitude/speed',  `height of foot lift', `heel strike during walking', `turning', and `arm swing') by observation. \texttt{MDS-UPDRS} item 3.10 is scored on a 5-level scale that assesses the severity of PD gait impairment, ranging from a score of 0 indicating no motor impairments to a score of 4 for patients unable to move independently (see Fig.~\ref{fig:pose_progression}). 

We propose a method based on videos to assess PD severity related to gait and posture impairments. Although there exist a few video-based methods which assess gait for PD diagnosis \cite{cho2009vision,xue2018vision,han2006gait}, we define a new task and a principled benchmark by estimating the standard \texttt{MDS-UPDRS} scores. There are several challenges to this new setting: (1) there are no baselines to build upon; (2) since it is harder to recruit patients with severe impairments, the number of participants in our dataset is imbalanced across \texttt{MDS-UPDRS} classes; (3) clinical datasets are typically limited in the number of participants, presenting difficulty for training deep learning models; (4) estimating \texttt{MDS-UPDRS} scores defines a multi-class classification problem on a scale of scores from 0 to 4, while prior work only focused on diagnosing PD \vs~normal. To address these challenges, our 3D pose estimation models are trained on large public datasets. Then, we use the trained models to extract 3D poses (3D coordinates of body joints) from our clinical data. Therefore, estimation of the \texttt{MDS-UPDRS} scores is only performed on low-dimensional pose data which are agnostic to the clinical environment and video background. 

To deal with data imbalance, we propose a model with a focal loss \cite{lin2017focal}, which is coupled with an ordinal loss component \cite{rennie2005loss} to leverage the order of the \texttt{MDS-UPDRS} scores. 

Our novel approach for automatic vision-based evaluation of PD motor impairments takes monocular videos of the \texttt{MDS-UPDRS} gait exam as input and automatically estimates each participants's gait score on the \texttt{MDS-UPDRS} standard scale. To this end, we first identify and track the participant in the video. Then, we extract the 3D skeleton (a.k.a. pose) from each video frame (visualized in Fig.~\ref{fig:pose_progression}). Finally, we train our novel temporal convolutional neural network (TCNN) on the sequence of 3D poses by training a Double-Features Double-Motion Network \cite{yang2019make} (DD-Net) with the new hybrid ordinal-focal objective, which we will refer to as hybrid Ordinal Focal DDNet (OF-DDNet) (see Fig. \ref{fig:pipeline}). 

The novelties of our work are three-fold: (1) we define a new benchmark for PD motor severity assessment based on video recordings of \texttt{MDS-UPDRS} exams; (2) for the first time, we propose a framework based on 3D pose acquired from non-intrusive monocular videos to quantify movements in 3D space; (3) we propose a method with a hybrid ordinal-focal objective that accounts for the imbalanced nature of clinical datasets and leverages the ordinality \texttt{MDS-UPDRS} scores. 

\section{Method}

\begin{figure}[t]
  \centering
  \includegraphics[width=\linewidth]{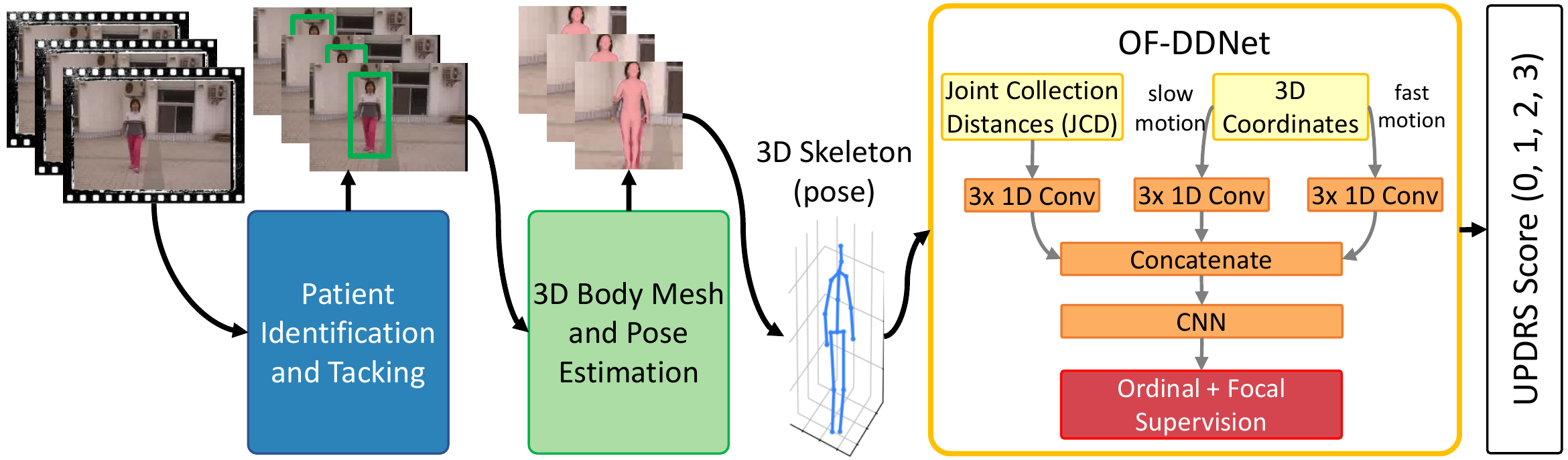}
  \caption{The proposed framework: we first track the participant throughout the video and remove other persons, \eg clinicians. Then, we extract the identified participants' 3D body mesh and subsequently the skeletons. Finally, our proposed OF-DDNet estimates the \texttt{MDS-UPDRS} gait score based on only the 3D pose sequence.}
  \label{fig:pipeline}
  \vspace{-2pt}
\end{figure}

As shown in Fig. \ref{fig:pipeline}, the input consists of a monocular video of each participant walking in the scene. First, we track each participant in the video using the SORT (Simple Online and Realtime Tracking) algorithm \cite{bewley2016simple} and identify the bounding boxes corresponding to the participant. These bounding boxes along with the \texttt{MDS-UPDRS} exam video are passed to a trained 3D pose extraction model (denoted by SPIN) \cite{kolotouros2019learning}, which provides pose input to OF-DDNet. 

\subsection{Participant Detection and Tracking} 
We first detect and track the participant since videos may contain multiple other people, such as clinicians and nurses. To do this, we track each participant in the video with SORT, a realtime tracking algorithm for 2D multiple object tracking in video sequences \cite{yang2019make}. SORT uses a  Faster Region CNN (FrRCNN) as a detection framework \cite{darknet13}, 
a Kalman filter \cite{kalman1960new} as the motion prediction component, and  
the Hungarian algorithm \cite{kuhn1955hungarian} for matching the detected boxes.

The participant is assumed to be in all frames, hence we pick the tracked person who is consistently present in all frames with the greatest number of bounding boxes as the patient. 

\subsection{3D Body Mesh and Pose Extraction}
Next, we extract the 3D pose from each frame by feeding the corresponding image and the bounding box found in the previous step as input to SPIN (SMPL oPtimization IN the loop) \cite{kolotouros2019learning}. SPIN is a state-of-the-art neural method for estimating 3D human pose and shape from 2D monocular images. Based on a single 2D image, the Human Mesh Recovery (HMR) regressor provided by \cite{kanazawa2018end} generates predictions for pose parameters $\theta_{reg}$, shape parameters $\beta_{reg}$, camera parameters $\Pi_{reg}$, 3D joints $X_{reg}$ of the mesh and their 2D projection $J_{reg} = \Pi_{reg}(X_{reg})$. Following the optimization routine proposed in SMPLify \cite{bogo2016keep}, these are initial parameters for the SMPL body model \cite{loper2015smpl}, a function $M(\theta, \beta)$ of pose parameters $\theta$ and shape parameters $\beta$ that returns the body mesh. A linear regressor $W$ performs regression on the mesh to find 3D joints $J_{smpl}$. These regressed joint values are supplied to the iterative fitting routine, which encourages the 2D projection of the SMPL joints $J_{smpl}$ to align with the annotated 2D keypoints $J_{reg}$ by penalizing their weighted distance. The fitted model subsequently provides supervision for the regressor, forming an iterative training loop. In our proposed method, we generate 3D pose for each video frame by performing regression on the 3D mesh output from SMPL, which has been fine-tuned in the SPIN loop. SPIN was initialized with pretrained SMPL \cite{loper2015smpl} and HMR pretrained on the large Human3.6M \cite{ionescu2013human3} and MPI-INF-3DHP \cite{VNect_SIGGRAPH2017} datasets, providing over 150k training images with 3D joint annotations, as well as large-scale datasets with 2D annotations (\eg COCO \cite{lin2017focal} and MPII \cite{andriluka20142d}).

\subsection{Gait Score Estimation with OF-DDNet}
Our score estimation model, OF-DDNet, builds on top of DD-Net \cite{yang2019make} by adding a hybrid ordinal-focal objective. DD-Net \cite{yang2019make} was chosen for its state-of-the-art performance at orders of magnitude smaller in parameter size than comparable methods. OF-DDNet takes as input 3D joints and outputs the participant's \texttt{MDS-UPDRS} gait score. Our model has a lightweight TCNN-based architecture that prevents overfitting. To address the variance of 3D Cartesian joints to both location and viewpoint, two new features are calculated: (1) Joint Collection Distances (JCD) and (2) two-scale motion features. JCD is a location-viewpoint invariant feature that represents the Euclidean distances between joints as a matrix $M$, where $M_{ij}^k = \|J_i^k-J_j^k\|$ for joints $J_i$ and $J_j$ at frame $k$ of total $K$ frames. Since this is a symmetric matrix, only the upper triangular matrix is preserved and flattened to a dimension of ${n \choose 2}$ for $n$ joints. A two-scale motion feature is introduced for global scale invariance which measures temporal difference between nearby frames. To capture varying scales of global motion, we calculate slow motion ($M_k^{slow}$) and fast motion ($M_k^{fast}$)
\begin{equation}
\begin{aligned}
M_k^{slow} &=S^{k+1} - S_k, k\in \{1,2,3,...,K-1\} , \\
M_k^{fast} &=S^{k+2} - S_k, k\in \{1,3, 5,...,K-2\},
\end{aligned}
\end{equation}
where $S_k = \{J_1^k, J_2^k, ...J_{n}^k\}$ denotes the set of joints for the $k^\text{th}$ frame. The JCD and two-scale motion features are embedded into latent vectors at each frame through a series of convolutions to learn joint correlation and reduce the effect of skeleton noise. Then, the embeddings are concatenated and run through a series of 1D convolutions and pooling layers, culminating with a softmax activation on the final layer to output a probability distribution for each class. 

\subsection{Hybrid Ordinal-Focal Loss}

To leverage the ordinal nature of \texttt{MDS-UPDRS} scores and to combat the natural class imbalance in clinical datasets, we propose a hybrid ordinal ($\mathcal{O}$) focal ($\mathcal{F}$) loss with a trade-off hyperparamter $\lambda$ as $
  \mathcal{L}=\mathcal{F}+\lambda\mathcal{O}.
$
Although many regression or threshold-based ordinal loss functions exist \cite{rennie2005loss,pang2017deeprank}, this construction allows its use in conjunction with our focal loss. 

\vspace{3pt}\noindent\textit{Focal Loss} is introduced to combat class imbalance \cite{lin2017focal}. It was initially proposed for binary classification, but it is naturally extensible to multi-class classification (\eg $C>2$ classes). We apply focal loss for predicting label $y$ with probability $p$:
\begin{equation}
\mathcal{F}(y, p) = \sum_{i=1}^C -\alpha(1 - p_i)^\gamma y_i log(p_i).
\end{equation}
The modulating factor $(1-p_i)^\gamma$ is small for easy negatives where the model has high certainty and close to 1 for misclassified examples. This combats class imbalance by down-weighting learning for easy negatives, while preserving basic cross-entropy loss for misclassified examples. We set the default focusing parameter of $\gamma = 2$ and weighting factor $\alpha=0.25$ as suggested by \cite{lin2017focal}. 

\vspace{3pt}\noindent\textit{Ordinal Loss} is used to leverage the intrinsic order in the \texttt{MDS-UPDRS} scores. We implement a loss function that penalizes predictions more if they are violating the order. This penalization incorporates the actual labels $\bar{y}\in\{0,1,2,3\}$ to indicate order instead of the probability vectors used in cross-entropy. Given the estimated label $\hat{\bar{y}}\in\{0,1,2,3\}$, we calculate the absolute distance $w = |\bar{y}-\hat{\bar{y}}|$ and incorporate this with categorical cross-entropy to generate our ordinal loss: 
\begin{equation}
    \mathcal{O}(y,p) = -\frac{1 + w}{C} \sum_{i=1}^Cy_ilog(p_i).
\end{equation}

\section{Experiments}

\subsection{Dataset}
We collected video recordings from 30 research participants who met UK Brain Bank diagnostic criteria of \texttt{MDS-UPDRS} exams scored by a board-certified movement disorders neurologist. All videos of PD participants were recorded during the off-medication state, defined according to previously published protocols \cite{poston2016compensatory}. All study procedures were approved by the Stanford Institutional Review Board and written informed consent was obtained from all participants in this study. 

We first extracted the sections of the video documenting the gait examination, in which participants were instructed to walk directly toward and away from the camera twice. The gait clips range from 17 seconds to 54 seconds with 30 frames per second.

Our dataset includes 21 exams with score 1, 4 exams with score of 2, 4 exams with score of 3 and 1 exam with score 0. Participants who cannot walk at all or without assistance from another person are scored 4, thus we exclude this class from our analysis due to the difficulty in obtaining videos recordings of their gait exam.

To augment the normal control cohort (\ie score 0), we include samples from the publicly available CASIA Gait Database A \cite{wang2003silhouette}, a similar dataset with videos of 20 non-PD human participants filmed from different angles. 

We extracted corresponding videos where participants walk directly toward and away from the camera, with length of minimum 16 and maximum 53 seconds. The underlying differences between the datasets should not bias our analyses because all score estimation algorithms operate on pose data with similar characteristics (same view points and duration) across all classes and we normalize and center the pose per participant by aligning temporal poses based on their hip joint. 

\subsection{Setup}
We preprocess our dataset by 1) clipping each video into samples of 200 frames each, where the number of clips per exam depends on its length, 2) supplying two additional cropped videos per exam for sparse classes 2 and 3 and 3) joint normalization and centering at the mid-hip. To address the subjective nature of \texttt{MDS-UPDRS} scoring by clinicians, we incorporate a voting mechanism. Each sub-clip is labeled same as the exam itself for training to independently examine each sub-part of the exam. This voting mechanism adds robustness to the overall system and allows us to augment the dataset for proper training of the TCNN. To account for the limited dataset size, all evaluations in this study were performed using a participant-based leave-one-out cross-fold validation on all 50 samples. We note that the clips and crops for each exam are \textit{never} separated by the train/test split. Optimal hyperparameters for the gait scoring model were obtained by performing a grid search using inner leave-one-out cross validation and the Adam optimizer $(\beta_1 = 0.9, \beta_2 = 0.999)$ \cite{kingma2014adam}. Best performance was achieved at 600 epochs, batch size of 64, filter size of 32 and an annealing learning rate from $1^{-3}$ to $1^{-6}$. For evaluation, we report per-class and macro average $F_1$, area under ROC curve (AUC), precision (Pre), and recall (Rec). 

\begin{table}[t]
\begin{minipage}[b]{.48\textwidth }%
  \caption{Per-class \texttt{MDS-UPDRS} gait score prediction performance of our method.   \vspace{6pt}}

  \label{ref:optimal}
  \centering
  \begin{tabular}{lcccc}
    \toprule
    \textbf{Gait Score}     & $F_1$     &   \textbf{AUC}      & \textbf{Pre}       & \textbf{Rec} \\
    \midrule
 0              & 0.91    & 0.93    & 0.91   & 0.91\\
 1              & 0.81    & 0.91    & 0.73   & 0.91\\
 2              & 0.73    & 0.87    & 0.80   & 0.67\\
 3              & 0.86    & 0.90    & 1.00   & 0.75\\

    \midrule
    Macro Average              & 0.83    & 0.90    & 0.86   & 0.81\\
    \bottomrule
  \end{tabular}
\hrule height 0pt
\end{minipage}%
~~
\begin{minipage}[b]{.48\textwidth}
  \centering
  \includegraphics[width=\textwidth]{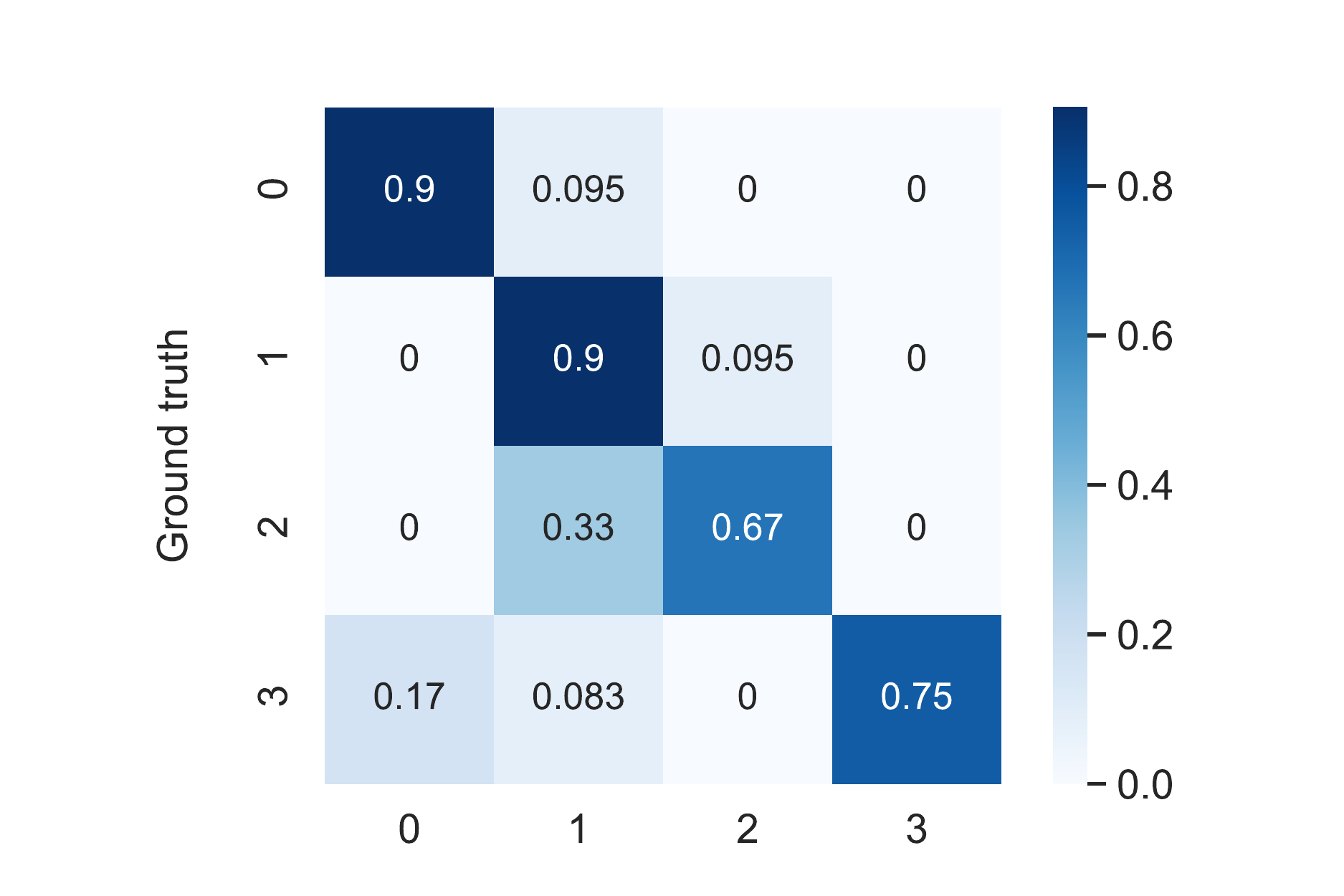}
  \captionof{figure}{\strut Confusion matrix of OF-DDNet. \label{fig:cm}}
\hrule height -12pt
\end{minipage}
\end{table}

\subsection{Baseline Methods and Ablation Studies}
We compare our results with several baselines: 1) we feed raw 3D joints from SPIN directly into a 1D CNN modeled after DD-Net architecture sans double features and embedding layer (see Fig.~\ref{fig:pipeline}), 2) OF-CNN, the same as (1) but with our OF loss, and 3) the original DD-Net \cite{yang2019make} with basic cross-entropy loss. We also conduct an ablation study on the choice of pose extraction method by 4) using 2D joints (instead of 3D) extracted with OpenPose \cite{cao2017realtime} as input to OF-DDNet. 
To evaluate the hybrid loss function, we separately examine our method 5) without the focal loss component and 6) without the ordinal component. 
We further examine our ordinal component by replacing it with 7) a regression loss (MSE) for DD-Net with an extra sigmoid-activated dense output layer and finally with 8) DeepRank \cite{pang2017deeprank}, a ranking CNN which cannot be combined with focal loss. 

\subsection{Results}

\begin{table}[t]
\setlength\tabcolsep{3.5pt} 
\caption{Comparison with baseline and ablated methods. * indicates statistical difference at $(p<0.05)$ compared with our method, measured by the Wilcoxon signed rank test \cite{wilcoxon1992individual}. Best results are in bold. See text for details about compared methods. }
\label{ref:big-table}
\centering
{\scriptsize
\begin{tabular}{lcccc|lcccc}
\toprule
 \textbf{Method} & $F_1$ & \textbf{AUC} & \textbf{Pre} & \textbf{Rec} &  \textbf{Method} & $F_1$ & \textbf{AUC} & \textbf{Pre} & \textbf{Rec}\\
\midrule
OF-DDNet (Ours)   & \textbf{0.83}    & \textbf{0.90}    & \textbf{0.86}   & \textbf{0.81}&
5) Ours w/o focal & 0.79 & 0.83 & 0.83  &    0.76\\

1) Baseline CNN$^\ast$ & 0.73 & 0.86 & 0.79   &   0.69  &    
6) Ours w/o ordinal & 0.78 & 0.88 & 0.84   &   0.74 \\

2) Baseline OF-CNN$^\ast$     &  0.74    & 0.83    & 0.79   & 0.71& 
7) Regression$^\ast$ & 0.67&n/a& 0.70 & 0.65 \\

3) DD-Net$^\ast$  \cite{yang2019make} & 0.74 & 0.84 & 0.80 & 0.69 & 
8) DeepRank$^*$ \cite{pang2017deeprank} &0.74& 0.80& 0.79& 0.71\\

4) 2D joints$^\ast$ \cite{cao2017realtime}        &  0.61    & 0.77    & 0.61   & 0.62 & \\
\bottomrule
\end{tabular}}
\end{table}
The results of our proposed OF-DDNet are summarized in Table~\ref{ref:optimal}. Our method sets a new benchmark for this task with macro-average $F_1$-score of 0.83, AUC of 0.90, precision of 0.86, and balanced accuracy (average recall) of 81\%. As seen in the confusion matrix (Fig.~\ref{fig:cm}), the overall metrics for well-represented classes control and class 1 are fairly high, followed by class 3 and then class 2. We observe that class 2 is strictly misclassified as lower severity. The results of comparisons with baseline and ablated methods are summarized in Table~\ref{ref:big-table}. Our proposed method achieves significantly better performance than many other methods based on the Wilcoxon signed rank test \cite{wilcoxon1992individual} (p < 0.05), and consistently outperforms all other methods.

Our results show that all methods have higher performance on 3D joints input than 2D input, as even a baseline 1D CNN has better performance than the full DD-Net model with 2D joints. This demonstrates that 3D joints provide valuable information for the prediction model, which has not been explored before. Similarly, we note that on 3D joint input, all classification methods outperformed the regression model, suggesting that classification outperforms regression at this task. Regarding the loss function, OF-DDNet significantly outperforms our baseline CNN with categorical cross-entropy. Adding ordinal (Method 5 in the Table) and focal (Method 6) losses to baseline DD-Net both improve accuracy, but their combined performance (OF-DDNet) outperforms all. DeepRank (Method 7) had high confidence on predictions and poor performance on sparse classes, suggesting an overfitting problem that encourages the use of a simple ordinal loss for our small dataset. 

\section{Discussion}
Our method achieves compelling results on automatic vision-based assessment of PD severity and sets a benchmark for this task. We demonstrate the possibility of predicting PD motor severity using only joint data as the input to a prediction model, and the efficacy of 3D joint data in particular. Furthermore, we show the effectiveness of a hybrid ordinal-focal loss for tempering the effects of a small, imbalanced dataset and leveraging the ordinal nature of the \texttt{MDS-UPDRS}. However, it is necessary to note that there is inherent subjectivity in the \texttt{MDS-UPDRS} scale \cite{evers2019measuring} despite attempts to standardize the exam through objective criterion (\eg stride amplitude/speed, heel strike, arm swing). Physicians often disagree on ambiguous cases and lean toward one score versus another based on subtle cues. Clinical context suggests our results are consistent with physician experience. As corroborated in the results of OF-DDNet, the most difficult class to categorize in clinical practice is score 2 since the \texttt{MDS-UPDRS} defines its distinction from score 1 solely by ``minor'' versus ``substantial'' gait impairment, shown in Fig.~\ref{fig:pose_progression}. Control (class 0) exhibits high arm swing and range of pedal motion while classes 1 and 2 have progressively reduced mobility and increased stiffness (\ie reduced arm swing and stride amplitude/foot lift). Class 3 exhibits high imbalance issues with stooped posture and lack of arm swing, which aids mobility, presenting a high fall risk. In practice, class 3 is easier to distinguish from the other classes because it only requires identifying that a participant requires an assisted-walking device and cannot walk independently . Likewise, our model performs well for class 3 except in challenging cases which may require human judgement, such as determining what constitutes ``safe'' walking. 

This study presents a few notable limitations. A relatively small dataset carries risk of overfitting and uncertainty in the results. We mitigated the former through data augmentation techniques and using simple models (DD-Net) instead of deep or complex network architectures; and the latter with leave-one-out cross validation instead of the traditional train/validation/test split used in deep learning. Similarly, our classes are imbalanced with considerably fewer examples in classes 2 and 3 than in classes 0 and 1, which we attempt to address through our custom ordinal focal loss and by augmenting sparse classes through cropping. Additionally, due to a shortage of control participants in our clinical dataset, we include examples of non-PD gait from the public CASIA dataset. The data is obfuscated by converting to normalized pose, which has similar characteristics across both datasets. However, expanding the clinical dataset by recruiting more participants from underrepresented classes would strengthen the results and presents a direction for future work. 

\section{Conclusion}
In this paper, we presented a proof-of-concept of the potential to assess PD severity from videos of gait using an automatic vision-based approach. We provide a first benchmark for estimating \texttt{MDS-UPDRS} scores with a neural model trained on 3D joint data extracted from video. This method works even with a small dataset due to data augmentation, the use of a simple model and our hybrid ordinal focal loss and has opportunity for application to similar video classification problems in the medical space. Our proposed method is simple to set up and use because it only requires a video of gait as input; thus, in remote or resource-limited regions with few experts it provides a way to form estimates of disease progression. In addition, such scalable automatic vision-based methods can help perform time-intensive and tedious collection and labelling of data for research and clinical trials. In conclusion, our work demonstrates how computer-assisted intervention (CAI) technologies can provide clinical value by reliably and unobtrusively assisting physicians by automatic monitoring of PD patients and their motor impairments. 

\section*{Acknowledgment} 
This research was supported in part by NIH grants AA010723, AA017347, and AG047366. This study also benefited from Stanford Institute for Human-centered Artificial Intelligence (HAI) AWS Cloud Credit.

\bibliographystyle{splncs03}
{ 
\bibliography{ref}
}

\end{document}